\documentclass[11pt,a4paper]{article}
\usepackage[hyperref]{acl2021}
\usepackage{times}
\usepackage{latexsym}
\usepackage{bm}
\usepackage{amsfonts}
\usepackage{balance}

\usepackage{microtype}

\usepackage{graphicx}
\usepackage{float}
\usepackage{subfigure}
\usepackage{amsmath}
\aclfinalcopy % Uncomment this line for the final submission
\def\aclpaperid{309} %  Enter the acl Paper ID here

\newcommand{\cparagraph}[1]{\vspace{1.5mm}\noindent\textbf{#1}}
\newcommand*\samethanks[1][\value{footnote}]{\footnotemark[#1]}

\title{Maria: A Visual Experience Powered Conversational Agent}

\author{Zujie Liang$^{1}$\thanks{\ \ Work performed during the internship at Microsoft.}
{}
\thanks{\ \ Equal contribution.} \quad
Huang Hu$^2$\samethanks \quad Can Xu$^2$ \quad Chongyang Tao$^{2}$ \quad
	\\\textbf{Xiubo Geng$^2$ \quad Yining Chen$^2$ \quad Fan Liang$^1$ \quad Daxin Jiang$^2$\thanks{\ \ Corresponding author.}}\\
	$^1$School of Electronics and Information Technology,\\ Sun Yat-sen University, Guangzhou, China\\
	$^2$Microsoft STCA NLP Group, Beijing, China \\
	{ \tt $^1$\{liangzj9@mail2.sysu.edu.cn, isslf@mail.sysu.edu.cn\}} \\
	{ \tt $^2$\{huahu,caxu,chotao,xigeng,yinichen,djiang\}@microsoft.com} \\}
\date{}

\begin{document}
\maketitle

\begin{abstract}

Arguably, the visual perception of conversational agents to the physical world is a key way for them to exhibit the human-like intelligence.
Image-grounded conversation is thus proposed to address this challenge.
Existing works focus on exploring the multimodal dialog models that ground the conversation on a given image.
In this paper, we take a step further to study image-grounded conversation under a fully open-ended setting where no paired dialog and image are assumed available.
Specifically, we present Maria, a neural conversation agent powered by the visual world experiences which are retrieved from a large-scale image index. 
Maria consists of three flexible components, \textit{i.e.}, text-to-image retriever, visual concept detector and visual-knowledge-grounded response generator. 
The retriever aims to retrieve a correlated image to the dialog from an image index, while the visual concept detector extracts rich visual knowledge from the image.
Then, the response generator is grounded on the extracted visual knowledge and dialog context to generate the target response. 
Extensive experiments demonstrate Maria outperforms previous state-of-the-art methods on automatic metrics and human evaluation, and can generate informative responses that have some visual commonsense of the physical world.\footnote{The dataset and code are publicly available at \href{https://github.com/jokieleung/Maria}{https://github.com/jokieleung/Maria}} 

\end{abstract}
\section{Introduction}
Building intelligent conversational agents that can not only converse freely with human but also have the ability to perceive the physical world, has been one of the longest standing goals of natural language processing (NLP) and artificial intelligence (AI).
Although the recent large-scale conversation models trained on text-only corpora, such as Meena \cite{adiwardana2020towards}, Blender \cite{roller2020recipes} and DialoGPT \cite{zhang2019dialogpt}, have shown the compelling performance, they are still lack of the perception ability to our physical world.
A recent study \cite{bisk2020experience} points out the successful linguistic communication relies on a shared experience of the world that makes language really meaningful.
The visual perception is a rich signal for modeling a vastness of experiences in the world that cannot be documented by text alone \citep{harnad1990symbol}.
\begin{figure}[!t]
	\centering
	\includegraphics[width=0.46\textwidth]{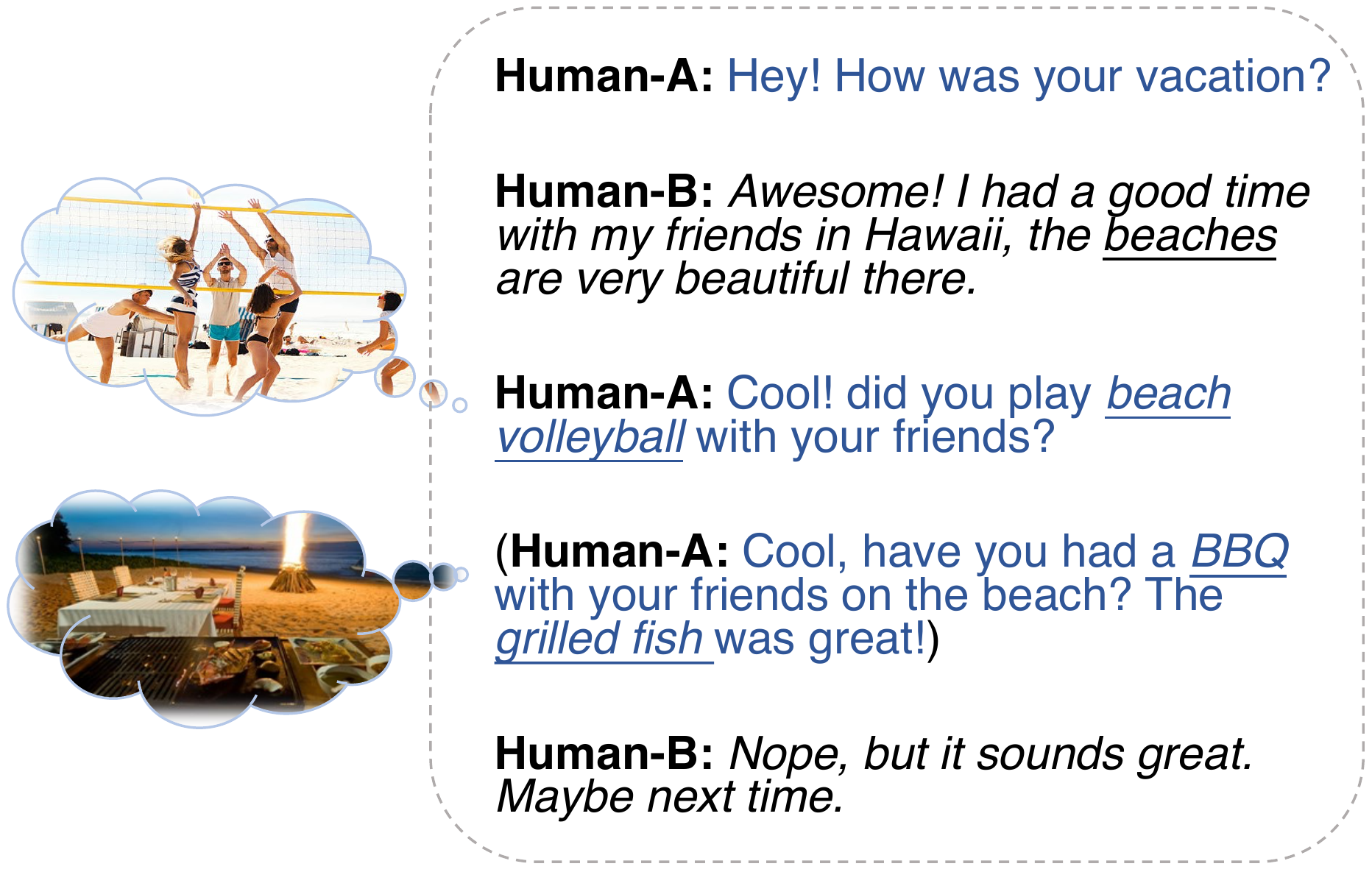}
	\vspace{-0.5em}
	\caption{An example of human conversations. When human-B talks about vacation on the beach of Hawaii, human-A recalls his/her past experience of playing volleyball or having BBQ on the beach.}
	\vspace{-1em}
	\label{fig:dialog_example}
\end{figure}
On the other hand, human-human conversations involve their understandings of context, the background knowledge they had, and perhaps most importantly the experiences of the world they shared, \textit{e.g.}, what they have seen before.

\begin{figure*}[!t]
	\centering
	\includegraphics[width=0.95\textwidth]{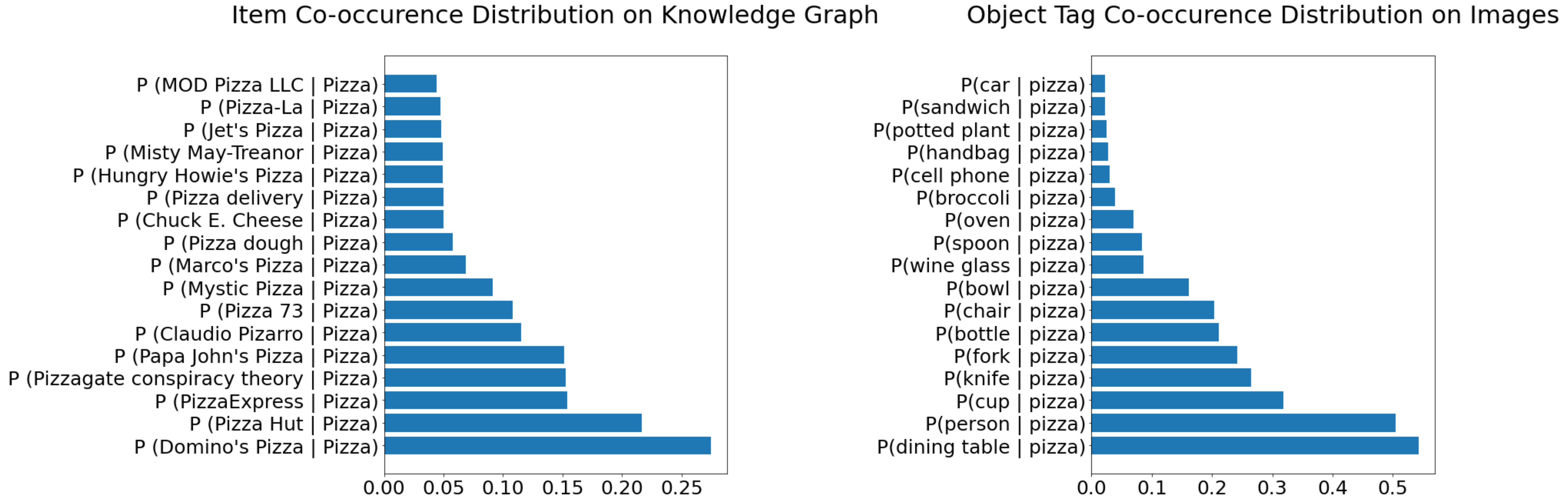}
	\vspace{-0.5em}
	\caption{The word co-occurrence distribution with ``pizza'' on Google knowledge graph and MS-COCO images.}
	\vspace{-1em}
	\label{fig:tag_distribution}
\end{figure*}

Figure~\ref{fig:dialog_example} shows a conversation between humans. Human-A recalls his/her past experience of playing volleyball or having BBQ on the beach when human-B talks about vacation on the beach of Hawaii. 
However, the association relationship between beach and volleyball (or BBQ) is hard to capture in traditional knowledge bases, such as knowledge graph.
Motivated by this, we select a common word ``pizza'' and collect the top 17 words that mostly co-occur with ``pizza'' on Google Knowledge Graph\footnote{\href{https://developers.google.com/knowledge-graph/}{https://developers.google.com/knowledge-graph/}} and MS-COCO images\footnote{We calculate the co-occurrence distribution of object tags from the images in MS-COCO dataset. More examples could be found in Appendices.} \cite{lin2014microsoft}. 
As shown in Figure~\ref{fig:tag_distribution}, the words co-occurring with ``pizza'' on knowledge graph tend to be the abstract concepts, while the co-occurrence relationship of object tags on images reflects some commonsense of our physical world, \textit{e.g.}, ``pizza'' is usually on the ``dining table'', people usually use ``knife'' when eating ``pizza''.
Interestingly, we found the ``pizza'' also co-occurs with ``cell phone'' and even ``plotted plant''.
This indicates when people eat pizza, they sometimes would put their cell phones aside on the table, or there might exist some plotted plants in the restaurant.
Thus, empowering conversational agents to have the visual perception ability about the physical world is a key way for them to exhibit the human-like intelligence.

The existing works \cite{mostafazadeh2017image,huber2018emotional,shuster2020image} focus on exploring the multimodal dialog models that ground the conversation on a given image. 
Recently, \citet{yang2020open} propose to learn the dialog generation model with both image-grounded dialogs and textual dialogs by resorting to text-to-image synthesis techniques \cite{xu2018attngan,qiao2019mirrorgan} to restore a latent image for the text-only dialog.
Even so, these works are still constrained by the assumption that the dialog is conducted center around a given (or synthesized) image.

In this paper, we take a step further to extend the assumption of image-grounded conversation to a fully open-ended setting where no image-dialog pairs are assumed available.
Specifically, we present Maria, a neural conversational agent powered by visual world experiences which are retrieved from a pre-built image index, \textit{e.g.}, the Open Images Dataset \cite{kuznetsova2018open}.
Maria consists of three components: text-to-image retriever, visual concept detector, and visual-knowledge-grounded response generator.
The retriever is responsible for retrieving a piece of visual world experiences, \textit{e.g.}, a correlated image to the dialog from an image index.
The visual concept detector utilizes the object detector from UpDown \cite{anderson2018bottom} to extract the regions features (\textit{i.e.}, bboxes) and the corresponding visual concepts (\textit{i.e.}, tags) from the retrieval images.
Hence, we can construct (\textit{bboxes}, \textit{tags}, \textit{context}, \textit{response}) 4-tuple as the training data.
Finally, these constructed 4-tuples are used to train the visual-knowledge-grounded response generator, which is built on the top of a multi-layer Transformer architecture \cite{vaswani2017attention}. 
To effectively inject the visual knowledge into the response generator, we carry out the Masked Concept Prediction and Visual Knowledge Bias besides the response generation objective.
The former aims to align the semantic representations between textual words and image regions, while the latter tries to provide more visual knowledge to facilitate the dialog generation. 
The experimental results on Reddit Conversation Corpus \cite{dziri2018augmenting} demonstrate that Maria significantly outperforms previous state-of-the-art methods, 
and can generate informative responses with visual commonsense of our physical world.

Overall, the contributions of this paper are summarized as follows: 
\begin{itemize}
\setlength{\itemsep}{0pt}
\setlength{\parsep}{0pt}
    \item We explore the task of image-grounded dialog generation under a fully open-ended setting where no specific image-dialog pairs are assumed available, \textit{i.e.,} zero-resource image-grounded conversation. 
    To the best of our knowledge, this is the first work to connect dialog corpus with the unpaired image data;
    \item We present Maria, a neural conversational agent consisting of three flexible components, which can effectively capture the visual commonsense from images and accordingly generate informative and vivid responses;
    \item Extensive experiments on the widely used Reddit Conversation Corpus are conducted to justify the effectiveness of Maria.
\end{itemize}
\section{Related Work}
\cparagraph{Vision and Language}
\indent
In the research of vision and language, various tasks have been extensively studied, such as image captioning \cite{vinyals2015show,lu2017knowing,hu2020vivo}, visual question answering \cite{antol2015vqa,anderson2018bottom}, visual dialog \cite{das2017visual,das2017learning}.
Popular benchmark datasets in this area include MS-COCO \cite{lin2014microsoft}, VisDial \cite{das2017visual} and Visual Genome \cite{krishna2017visual}.
Visual dialog is a task to answer the questions about the factual content of the image in a multi-turn manner.
Differently, image-grounded conversation studies how to reply to a dialog context and a given image with proper responses in an open-ended way.

\cparagraph{Dialog Generation}
\indent
Encouraged by the success of the neural sequence-to-sequence architecture \cite{sutskever2014sequence} on machine translation, end-to-end neural approaches on open-domain dialog generation \cite{vinyals2015neural,shang2015neural,serban2016building,sordoni2015neural,xing2017topic,wu2018neural,zhang2019dialogpt,xu2019neural,adiwardana2020towards} have been widely studied in literature.
Recently, there is an emerging trend towards grounding the dialog generation models on the external knowledge, such as knowledge graphs \cite{zhou2018commonsense}, documents \cite{ghazvininejad2018knowledge,dinan2018wizard,kim2020sequential,zhao2020low,zhao2020knowledge,li2020zero} and images \cite{mostafazadeh2017image,shuster2020image,yang2020open}.
Different from the previous work on knowledge-grounded conversation that connects dialogs with unpaired document knowledge~\cite{li2020zero}, our work lies in the research of image-grounded conversation where a response is generated with a dialog context and a given image. 
Existing works~\cite{mostafazadeh2017image,shuster2020image,yang2020open} in this direction assume there is a given (or synthesized) image for the dialog and explore the multimodal dialog models. 
In contrast to these works, we study the image-grounded conversation under a fully open-ended assumption where no paired dialog and image are assumed available, \textit{i.e.}, zero-resource image-grounded conversation.

\section{Problem Formalization}

Suppose we have a dialog set $\mathcal{D}=\left\{\left(C_{i}, R_{i}\right)\right\}_{i=1}^{n}$, where $\forall i \in\{1, \ldots, n\}$, $C_{i}$ refers to a dialog context and $R_{i}$ is a response to $C_{i}$. 
We assume there is a set of images $\mathcal{V}=\left\{V_{j}\right\}_{j=1}^{m}$, where $\forall j \in\{1, \ldots, m\}$, $V_{j}$ denotes an image.
$\forall C \in\mathcal{D}$, we assume that there is an image $V$ that triggered by the given dialog context $C$ and response $R$. 
Our goal is to estimate a generation model $P(R|V,C)$ from $\mathcal{D}$ and $\mathcal{V}$. 
Thus, given a new dialog context $C$ associated with an image $V$, the model can generate a response $R$ according to $P(R|V,C)$.

\begin{figure}[t]
	\centering
	\includegraphics[width=0.48\textwidth]{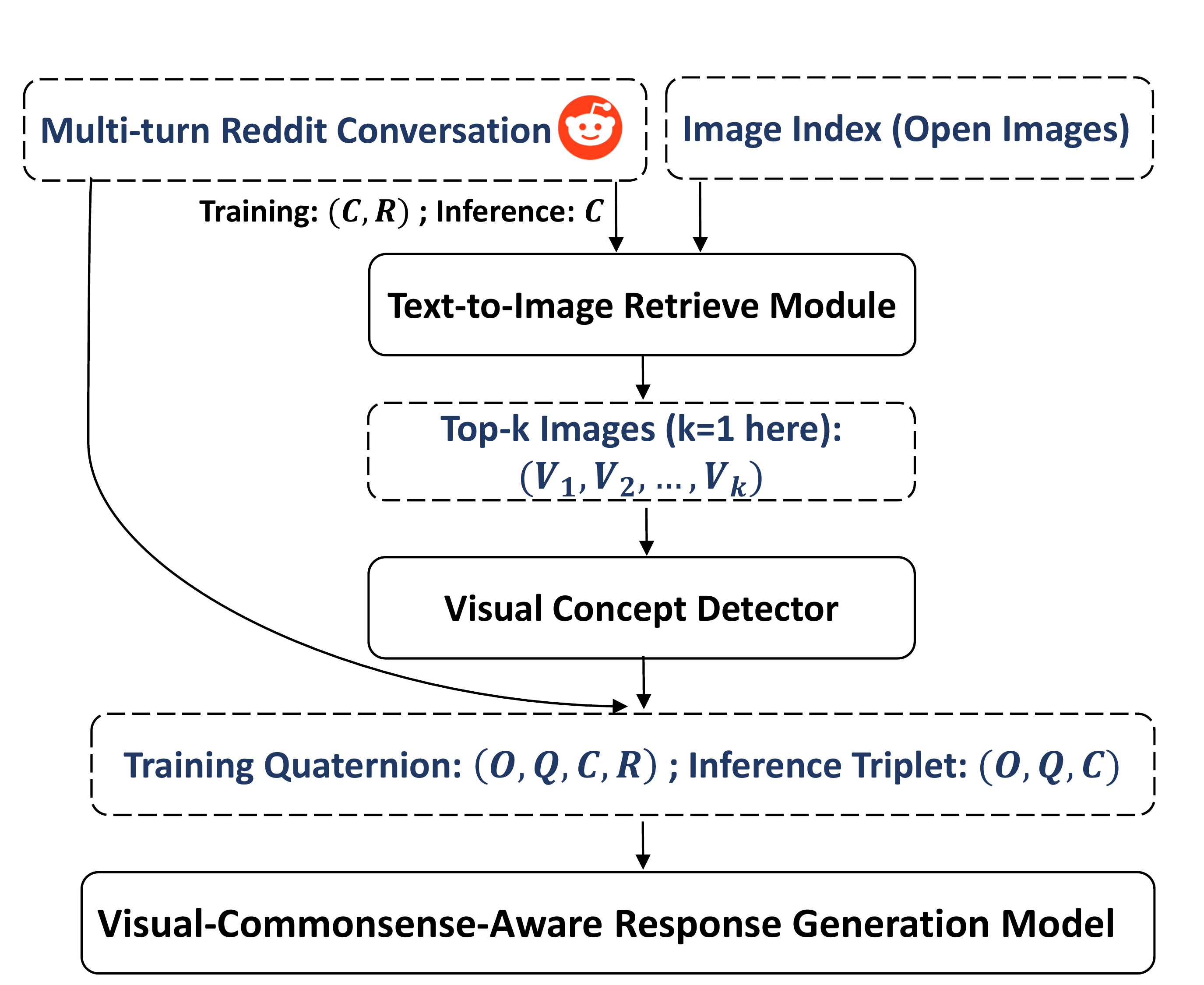}
	\vspace{-1.5em}
	\caption{The flowchart of our framework. $O,Q,C,R$ represents the image region features, extracted visual concepts, dialog context and response.}
	\vspace{-1em}
	\label{fig:framework_flowchart}
\end{figure}

\section{Methodology}
To learn such a generation model $P(R|V,C)$, we need to tackle several challenges: 
(1) How to bridge the gap between unpaired dialog corpus and image data;
(2) After obtaining the correlated images, how to extract the detailed visual features and concepts;
(3) How to effectively inject the visual knowledge into response generator and enable it to generate responses that are visual-knowledge-grounded.
Figure~\ref{fig:framework_flowchart} illustrates the framework of our approach. 
We first build a large-scale image dataset and leverage a cross-modal matching model to retrieve a correlated image using the content of the dialog.
Then an off-the-shelf object detector is applied to extracting the object features and visual concepts from the retrieval image. 
Finally, the response generator is trained to generate the target response conditioned on the context, extracted object features, and visual concepts.
In the rest of this section, we will elaborate these three modules.

\subsection{Text-to-Image Retriever}
In this section, we develop a retrieval model that assigns each dialog with a correlated image $V$. Specifically, we train a text-to-image matching model from image captioning dataset and utilize it to construct the $(C, R, V)$ triple data. 

\cparagraph{Modeling}
\indent 
To improve the efficiency of cross-modal retrieval model on large-scale dialog corpus and image dataset, we adopt a two-tower architecture \cite{lu2019vilbert} to accelerate the retrieval process where the image features can be pre-extracted offline.
The model takes a sentence $T$ and an image $V$ as input, and predicts the relevance score $s(T, V)$ between the sentence and the image. 
We use a text encoder and an image encoder to produce the representations of $T$ and $V$, respectively. 
The text encoder is a pre-trained BERT-base model \cite{devlin2018bert} and we use the hidden state of special token \texttt{[CLS]} as the embedding of $T$:
\begin{equation}
\label{eq:bert}
\boldsymbol{e_{t}}=BERT(T)
\end{equation}
Then a Multi-Layer Perceptron (MLP) projects the sentence embedding into the cross-modal space. We follow \citet{tan2020vokenization} to perform L2-normalization on the last output features, by which we can simplify the nearest neighbor search problem in the euclidean space to the Maximum Inner Product problem~\citep{mussmann2016learning}:
\begin{equation}
\label{eq:l2_norm}
\boldsymbol{{f}_{t}}\left(T\right)=\frac{{H}_{t}\left(\boldsymbol{e_{t}}\right)}{\left\|{H}_{t}\left(\boldsymbol{e_{t}}\right)\right\|}
\end{equation}
Similarly, the image encoder is composed of a pre-trained ResNeXt  backbone \cite{xie2017aggregated} and a MLP with L2 normalization:
\begin{equation}
\label{eq:image_encoder}
\boldsymbol{{f}_{v}}\left(V\right)=\frac{{H}_{v}\left(\boldsymbol{e_{v}}\right)}{\left\|{H}_{v}\left(\boldsymbol{e_{v}}\right)\right\|},\ 
\boldsymbol{e_{v}}=ResNeXt(V)
\end{equation}
Thus, we define the relevance score $s(T, V)$ as an inner product of the language feature representation ${f}_{t}\left(T\right)$ and image feature representation ${f}_{v}\left(V\right)$:
\begin{equation}
\label{eq:relevance_score}
s(T, V)=\boldsymbol{{f}_{t}}\left(T\right)^{\top} \boldsymbol{{f}_{v}}\left(V\right)
\end{equation}

\cparagraph{Training}
\indent We train the cross-modal matching model on MS-COCO image captioning dataset \cite{lin2014microsoft}, where each image is paired with 5 sentences describing its visual content. 
The model is optimized by minimizing the hinge loss so that the relevance score $s\left(T, V\right)$ of the positive image-sentence pair can be larger than the negative pair $s\left(T, V^{-}\right)$ by at least a margin $M$:
\begin{equation}
\label{eq:hinge_loss}
\begin{aligned}
&\mathcal{L}_{hinge}\left(T, V, V^{-}\right)= \\
&\sum_{i=1}^{l} \max \{0, M-s\left(T, V\right) \left.+s\left(T, V^{-}\right)\right\}
\end{aligned}
\end{equation}

\cparagraph{Inference}
\indent Given the trained retrieval model, we can now assign each dialog with a correlated image $V$.
To ensure the diversity and richness of the retrieval results, we fetch 500,000 images from the large-scale Open Images dataset \cite{kuznetsova2018open} as our image set $\mathcal{V}$. 
The image $V_{i} \in \mathcal{V}$ with the maximum relevance score is paired with the given dialog $(C_i, R_i) \in \mathcal{D}$.
Note that for the dialog in the training set, we use both the context $C$ and response $R$ are concatenated as the query for retrieval (\textit{i.e.}, $T=(C,R)$), which is beneficial to retrieving an image with the related visual knowledge. 
On the other hand, for the validation/test set of the dialog corpus, the query is only the context (\textit{i.e.}, $T=C$) so as to keep consistent with the real-world setting where the response is unavailable and need to be generated at inference. 

\subsection{Visual Concept Detector}

Given the correlated image $V_{i}$ to the dialog as the visual clue, we can now extract the visual knowledge from it. 
One naive approach is to utilize the CNN-based models to extract the latent image features. 
However, this approach does not consider the fine-grained representation modeling for images, which is crucial for the dialog model to understand the local visual features in images. 
To address this issue, we adopt an object detection model \cite{anderson2018bottom} pre-trained on Visual Genome \cite{krishna2017visual} to extract a set of salient object features  ${O}=\left\{\mathbf{o}_{k}\right\}_{k=1}^{K}$, where each object feature $\mathbf{o}_{k}$ is a 2048-dimensional vector. 
These features represent the images at the level of objects and other salient regions, which has proven to be vital in many high-level image understanding tasks. 
Besides, the same detector is used to extract a set of visual concepts ${Q}=\left\{{q}_{m}\right\}_{m=1}^{K}$, where each concept ${q}_{m}$ is the high-precision textual label of the visual region, \textit{e.g.}, “sunset", “melon", etc. 
In this manner, we simultaneously obtain the fine-grained image representations and the necessary visual concepts for the subsequent dialog generation.

\subsection{Visual-Knowledge-Grounded Response Generator}

In this section, we propose a unified architecture to effectively inject a set of region features and corresponding visual concepts into the response generation model.
In following parts, we describe the model design and training objectives in detail.

\begin{figure*}[!t]
	\centering
	\includegraphics[width=1\textwidth]{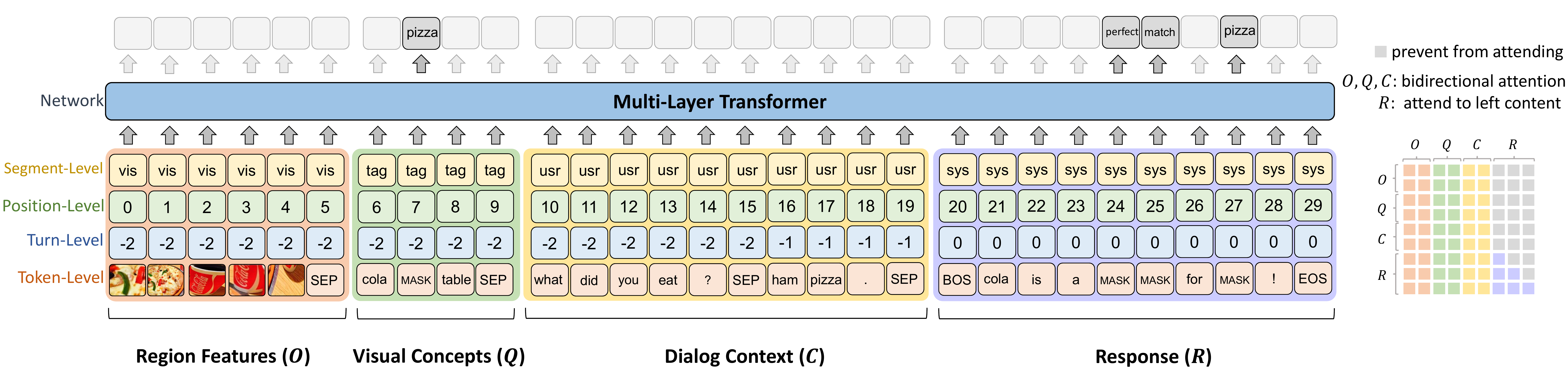}
	\vspace{-1.5em}
	\caption{The overview of the response generation model. There are four kinds of inputs, \textit{i.e.}, image region features $O$, extracted visual concepts $Q$, dialog context $C$ and response $R$. The self-attention mask in $R$ is unidirectional, \textit{i.e.}, can only attend to the left context, while the self-attention mask in other segments is bidirectional.}
	\vspace{-1em}
	\label{fig:model_overview}
\end{figure*}

\subsubsection{Model Architecture}

Figure~\ref{fig:model_overview} shows the architecture of our response generation model, which is a multi-layer transformer network for both bidirectional vision/context $(O,Q,C)$ encoding, and unidirectional response $R$ decoding, via the flexible self-attention masks inspired by ~\citep{unilm}. 

\subsubsection{Input Representation}

For each token, the final input representation to the multi-layer transformer network is the element-wise summation of four kinds of embeddings, including token-level, turn-level, position-level, and segment-level. Then, we concatenate all the input representations to one sequence for model training.

\cparagraph{Token-Level} 
\indent 
The token-level embeddings are the concatenation of $(O_{w},Q_{w},C_{w},R_{w})$, which denote the token embedding sequence of visual objects, visual concepts, contexts and response respectively.
Note that $O_{w}$ is the object embedding transformed by a linear layer into the same dimension as word embedding.
 
\cparagraph{Turn-Level}
\indent Since the dialog is multi-turn, we encode this turn order with a relative turn embedding~\citep{bao2020plato}. 
Specifically, the turn number is counted from the last utterance of the dialogue to the beginning.
Note that as for the tokens corresponding to $O$ and $Q$, we simply set them the same as the first utterance of $C$.

\cparagraph{Position-Level}
\indent Positional embedding encodes the signal of the token order in the total input sequence, which is the same as positional encoding of the original transformer \cite{vaswani2017attention}.
 
\cparagraph{Segment-Level}
\indent Segment embedding is employed to differentiate which segment the token is in, \textit{i.e.}, $O,Q,C$ or $R$.

\subsubsection{Masked Concept Prediction}

Due to the inherent gap between visual modality and textual modality, directly optimizing the model by response generation objective may result in the insufficient utilization of the visual knowledge. 
To align the semantic representations of two modalities, we devise Masked Concept Prediction (MCP) objective.
15$\%$ of the visual concepts are randomly replaced with \texttt{[MASK]} tokens in each training instance, which need to be predicted by the model.
However, one problem still remains, \textit{i.e.}, the visual concepts have no specific order when extracting from images.
In other words, we need to model MCP as a matching problem of set, which does not need to consider the order of predicted concepts when there are more than two concepts masked out simultaneously. 
To tackle this, inspired by \citet{hu2020vivo}, we adopt the Hungarian Matching Loss \cite{stewart2016end,carion2020end} to estimate an optimal mapping $\alpha$ so that the prediction for each masked position is assigned one of the target concepts. 
Here we denote the set of all input as $X=(O,Q,C,R)$,
the set of the bidirectional self-attention part of $X$ as $B=(O,Q,C)$,
the set of masked concepts as $\hat{Q}$,
the set of unmasked tokens as $B\backslash \hat{Q}$, 
and the prediction probabilities of the corresponding representations in the final layer of transformer as ${H}=\left\{h_{i}\right\}_{i=1}^{m}$ where $h_{i}$ is the probability distribution of the $i$-th masked position. 
Hence, the MCP loss can be defined as:
\begin{equation}
\label{L_MCP}
\begin{aligned}
&\mathcal{L}_{\mathrm{MCP}}({\hat{Q}}, {H}, \alpha)= \\
&-\sum_{q_{\alpha(i)} \in \hat{Q}}\log{h}_{i}\left(q_{\alpha(i)}\mid B \backslash \hat{Q}\right)  
\end{aligned}
\end{equation}
where $\alpha(i)$ is the index of the target concept assigned to the $i$-th prediction. 
When predicting a masked concept, the model will have to resort to visual region features, dialog contexts and other unmasked visual concepts. 
This would help the model to align the cross-modal representations between text and visual regions.

\subsubsection{Masked Response Prediction}

Encouraged by the success of UniLM~\citep{unilm} in Seq2Seq tasks, we adopt the Masked Response Prediction (MRP) objective to model the response generation. 
During training, 70$\%$ of the tokens in $R$ are randomly masked with the special token \texttt{[MASK]}. 
The model is optimized to recover the masked tokens. 
The masked response tokens and other unmasked tokens in the whole input sequence can be denoted as $\hat{R}$ and $X\backslash \hat{R}$, respectively. 
Suppose that $p_{i}$ is the conditional probability distribution of the $i$-th token in $R$, the MRP loss is the Negative Log-Likelihood (NLL) of the masked response tokens as follow:
\begin{equation}
\label{L_MRM}
\mathcal{L}_{\mathrm{MRP}}(X, \hat{R})=-\sum_{w_{i} \in \hat{R}} \log p_{i}\left(w_{i} \mid X \backslash \hat{R}\right)
\end{equation}
Note that the self-attention mask in $R$ is left-to-right, but the rest are bidirectional. 
In other words, the tokens in $O,Q$ and $C$ can attend to each other from both directions,
while the tokens in $R$ can attend all tokens in $O,Q,C$ and the leftward tokens in $R$ including itself.
MRP implicitly encourages the model to generate responses by learning the relationship among all input tokens. 

For decoding, we first encode the image regions, visual concepts, dialog contexts, and a special token \texttt{[BOS]} as input. 
Then the model starts the generation by feeding a \texttt{[MASK]} token and samples a word from the predicted distribution over vocabulary. 
Then, the \texttt{[MASK]} token is replaced by the generated token and a new \texttt{[MASK]} is appended to the input sequence for next word prediction. 
The generation process terminates when the model predicts \texttt{[EOS]} token or reaches the pre-defined maximum length.

\cparagraph{Visual Knowledge Bias}
\indent  Normally, the top projection layer of generation model produces a probability distribution over the vocabulary:
\begin{equation}
\bm{p} = softmax({W}{e^{r}}+{b}),
\end{equation}
where the ${e^{r}}\in \mathbb{R}^{d}$, ${W} \in \mathbb{R}^{|V| \times d}$ and ${b} \in \mathbb{R}^{|V|}$ are the last output of the transformer network, weight and bias parameters of the decoding head, respectively. 
$|V|$ denotes the vocabulary size. 
So far, the visual world knowledge is introduced into the response generation model by the shared-parameter self-attention layers.
To further inject the visual knowledge into the generation model, we design a simple but effective strategy, namely Visual Knowledge Bias (VKB). %Visual Vocabulary Bias. 
Concretely, an additional visual vocabulary bias $b_{q}$ is first calculated as follow: 
\begin{equation}
b_{q} = F_{q}({e^{q}_{avg}})
\end{equation}
where $F_{q}: \mathbb{R}^{d} \rightarrow \mathbb{R}^{|V|}$ is a projection layer. ${e^{q}_{avg}}$ denotes the average pooling on all hidden representations of visual concepts, \textit{i.e.}, ${e^{q}_{avg}}=AvgPooling({E^{q}})$ where ${E^{q}}=({e^{q}_{1}}, ..., {e^{q}_{K}})$. 
Then, we mask non-visual-concept tokens in the vocabulary and the masked vocabulary bias $\hat{b_{q}}\in \mathbb{R}^{|V|}$ is added to the top layer of generation model to get the final distribution over vocabulary:
\begin{equation}
    \hat{\boldsymbol{p}} = softmax(W{e^{r}}+b+\hat{b_{q}})
\end{equation}
We leverage this final vocabulary distribution to calculate the MRP loss in Eq.~\ref{L_MRM} to optimize the model. 
This visual knowledge bias would encourage the model to generate more visual knowledge related tokens in the response.

To sum up, the final objective of our response generation model is to minimize the integrated loss:
\begin{equation}
    \mathcal{L}=\mathcal{L}_{\mathrm{MRP}} + \mathcal{L}_{\mathrm{MCP}}
\end{equation}

\section{Experimental Setup}

\subsection{Datasets}
To evaluate the performance of Maria, we conduct comprehensive experiments on the Reddit dataset released by \citet{yang2020open}, which is a large-scale and high-quality multi-turn conversations extracted from Reddit Conversation Corpus \citep{dziri2019augmenting}.
Each dialog has 3 to 5 utterances, and the training/validation/test set has 1M/20K/20K dialogs respectively.

We train and validate the retrieval model using the Karpathy’s split\footnote{\href{https://cs.stanford.edu/people/karpathy/deepimagesent}{https://cs.stanford.edu/people/karpathy/deepimagesent}} of the MS-COCO image captioning data, where the images are split into 113.2K/5K/5K samples as training/validation/test set, respectively. 
After the retrieval model is trained, 
we fetch 500K images from the Open Images dataset as the image index,
and then retrieve images from it by dialog context and response to construct the training data for response generator. 

\subsection{Evaluation Metrics}
Both automatic metrics and human evaluation are employed to assess the performance of Maria and baselines. 
Automatic metrics include: 
(1) \textbf{Fluency}: perplexity (PPL) measures the confidence of the generated responses; 
(2) \textbf{Relevance}: BLEU-1~\citep{papineni2002bleu}, Rouge-L~\citep{lin2004rouge}, and we follow \citet{serban2017hierarchical} to utilize Embedding Average cosine similarity, Vector Extrema cosine similarity, and Embedding Greedy Matching score. All this metrics are calculated by running the public NLG evaluation script\footnote{\href{https://github.com/Maluuba/nlg-eval}{https://github.com/Maluuba/nlg-eval}}; 
(3) \textbf{Diversity}: Distinct-1 (Dist-1) and Distinct-2 (Dist-2)~\citep{li2016diversity} are defined as the number of distinct uni-grams or bi-grams divided by the total amount of words.

In human evaluation, we randomly select 100 dialogue contexts and the corresponding generated responses for Maria and compared baselines.
Three human annotators are asked to score the response quality on a scale of \{0, 1, 2\} from three aspects, including \textbf{Fluency}, \textbf{Relevance} and \textbf{Richness}. 
The higher score means the better. 
Since each response receives 3 scores on each aspect, we report the average scores over annotators and responses. 
The inter-annotator agreement is measured by Fleiss’ Kappa\citep{fleiss1973equivalence}.

\subsection{Implementation Details}
For the retrieval model, ResNeXt-101-32x8d feature is used as the visual embedding,
while the concatenation of the last 4 layers of BERT's outputs is used as the textual embedding. 
Both embeddings are then respectively fed into an MLP composed of three layers of size (1024, 1024, 512). 
When training the retrieval model, we set the margin $M=0.5$ for the hinge loss, and only tune the parameters of both MLPs while freezing the parameters of ResNeXt and BERT. 
The total training epoch is 20. 
At inference, the FAISS~\citep{johnson2019billion} library is utilized to accelerate the inner product search by batch processing. 
We use the off-the-shelf object detector from UpDown~\citep{anderson2018bottom} to extract top-k (k=36) image region features and the corresponding visual concepts. 
The detector is a Faster R-CNN~\citep{ren2015faster} model trained on the Visual Genome dataset \cite{krishna2017visual}.

For the response generation model, we set the number of transformer layers $L=12$ and the hidden embedding dimension $D=768$.
Besides, the network parameters are initialized by UniLM.
The maximum sequence lengths of context and response are set to 110 and 40, respectively. 
The sequence lengths of region features and concept tokens are both set to 36. 
The batch size is 64. 
We use the Adam Optimizer \cite{kingma2014adam} with a learning rate 3e-5 to train the response generation model.
The training is conducted on 4 Nvidia Tesla P40 24G GPU cards for 20 epochs. 

\subsection{Baselines}
We compare the following baselines in the experiments: 
(1) \textbf{Seq2Seq}: A standard Sequence to Seqence model with attention mechanism~\citep{bahdanau2014neural}. 
(2) \textbf{HRED}: A Hierarchical Recurrent Encoder-Decoder neural network~\citep{serban2016building}.
(3) \textbf{VHRED}: A variation of HRED that introduces latent variables into the generation~\citep{serban2017hierarchical}.
(4) \textbf{ReCoSa}: A hierarchical transformer-based model~\citep{zhang2019recosa} that achieves the state-of-the-art performance on benchmarks of dialog generation.
(5) \textbf{ImgVAE}: A dialog generation model~\citep{yang2020open} that is trained on both textual dialogs and image-grounded dialogs by recovering a latent image behind the textual dialog within a conditional variational auto-encoding framework.
(6) \textbf{DialoGPT}: An open-domain dialog model~\citep{zhang2019dialogpt} that fine-tunes GPT-2 \cite{radford2019language} on massive Reddit data. 
Since DialoGPT is a dialog generation model trained on the text-only corpus, we introduce it as an auxiliary baseline.
For a fair comparison, we choose the same model size ($L$=12,$D$=768) of DialoGPT (117M) as our model.

\begin{table*}[t]
\centering
\resizebox{150mm}{!}{
\begin{tabular}{l|cccccc|cc}
\hline
\textbf{Model}                & \textbf{PPL}   & \textbf{BLEU-1} & \textbf{Rouge-L} & \textbf{Average} & \textbf{Extrema} & \textbf{Greedy} & \textbf{Dist-1} & \textbf{Dist-2} \\ \hline
Seq2Seq \cite{bahdanau2014neural}                      & 77.27          & 12.21           & 10.81            & 78.38            & 40.06            & 62.64           & 0.53            & 1.96            \\
HRED \cite{serban2016building}                         & 84.02          & 11.68           & 11.29            & 75.54            & 37.49            & 60.41           & 0.89            & 3.21            \\
VHRED \cite{serban2017hierarchical}                         & 78.01          & 12.22           & 11.82            & 75.57            & 39.24            & 62.07           & 0.87            & 3.49            \\
ReCoSa \cite{zhang2019recosa}                       & 71.75          & 12.75           & 11.75            & 79.84            & 42.29            & 63.02           & 0.66            & 3.83            \\
ImgVAE \cite{yang2020open}                       & 72.06          & 12.58           & 12.05            & 79.95            & 42.38            & 63.55           & 1.52            & 6.34            \\ \hline
DialoGPT \cite{zhang2019dialogpt}                     & \textbf{36.03} & 5.87            & 5.20             & 77.80            & 35.40            & 58.39           & \textbf{10.41}           & \textbf{49.86}           \\ \hline
\textbf{Maria}   & \underline{54.38}          & \textbf{14.21}  & \textbf{13.02}   & \textbf{82.54}   & \textbf{44.14}   & \textbf{65.98}  & \underline{8.44}            & \underline{33.35}           \\ \hline
Maria (\textit{w/o} MCP)                & 66.71          & 13.91           & 11.60            & 81.59            & 41.06            & 64.10           & 8.36            & 31.80           \\
Maria (\textit{w/o} VKB)           &  65.51              & 12.76           & 11.76            & 82.49            & 40.22            & 64.49                & 7.15            & 29.44           \\
Maria (\textit{w/o} VKB \& MCP)         & 62.64          & 11.50           & 10.45            & 77.52            & 41.27            & 61.00           & 6.92            & 28.53           \\
Maria (\textit{w/o} images)             & 64.75          & 10.70           & 9.15             & 78.89            & 39.88            & 62.39           & 6.88            & 28.01           \\
Maria (\textit{w/o} concepts)           & 69.24          & 11.43           & 10.61            & \textbf{82.96}   & 41.02            & 65.07           & 4.56            & 16.44           \\
Maria (\textit{w/o} images \& concepts) & 69.50          & 10.75           & 8.34             & 80.62            & 41.15            & 64.25           & 3.69            & 10.11           \\ \hline
\end{tabular}
}
\caption{\label{tab:compare_with_sota}
Evaluation results of generated responses on the test set. Numbers in bold denote that the improvement over the best performing baseline is statistically significant. Numbers with underline refer to the best results except for the comparison to DialoGPT \cite{zhang2019dialogpt}.}
\vspace{-1.0em}
\end{table*}

\begin{table}[t]
\centering
\resizebox{77mm}{!}{
\begin{tabular}{lcccc}
\hline
Model          & \multicolumn{1}{c}{Fulency} & Relevance & Richness & Kappa \\ \hline
ImgVAE         &       1.79               &   0.58        &    0.67      &    0.67   \\
DialoGPT       &       \textbf{1.93}                      &   \underline{0.92}        &   \textbf{1.20}       &   0.59    \\
\textbf{Maria} &      \underline{1.89}                       &   \textbf{1.06}        &   \underline{0.97}       &  0.62     \\ \hline
\end{tabular}
}
\caption{\label{exp:human_eval}
Human evaluation results.
}
\vspace{-1.0em}
\end{table}

\section{Experimental Results}

\subsection{Automatic and Human Evaluations}
We summarize the experimental results of automatic evaluations in Table~\ref{tab:compare_with_sota}. 
Maria achieves the substantial performance improvements over baselines on all metrics except for the comparison to DialoGPT. 
Especially, Maria significantly surpasses ImgVAE on Dist-1/2, which indicates introducing richer visual knowledge, \textit{i.e.}, image region features and the corresponding visual concepts, is beneficial to generating more diverse and informative responses.
This also reflects in human evaluation of Table~\ref{exp:human_eval} that the richness score of Maria is higher than that of ImgVAE. 
Besides, in terms of relevance metrics including BLEU-1, Rouge-L, Average, Extrema and Greedy, 
Maria outperforms all baselines and even performs better than DialoGPT.
This indicates introducing the extra visual knowledge related to dialog context can further force the model to produce more relevant responses. 

On the other hand, the discrepancy of data distributions between the training data (\textit{i.e.}, Image-Chat~\citep{shuster2020image} dataset) and test data (\textit{i.e.}, Reddit conversation dataset) of the text-to-image synthesis model in ImgVAE limits its performance in practice.
Besides, constrained by the capability of the text-to-image synthesis model, 
the richness and diversity of the synthesized images are undesirable,
while Maria can retrieve a variety of images from the large-scale image index.
That may be the reason why ImgVAE consistently underperforms our Maria on relevance including automatic evaluation and human judgement, which also shows the superiority of the retrieval method for the zero-resource image-grounded conversation.  
Another observation is that Maria slightly underperforms DialoGPT on PPL and Dist-1/2. 
Since DialoGPT is a large-scale pre-training based dialog generation model and introduces the extra mutual information maximization objective to improve the informativeness of generated responses, which is consistent in human evaluation with respect to fluency and richness. 

\subsection{Ablation Study}
We conduct extensive ablation experiments over different model variants and input components to better understand their relative importance to the dialog generation task. 
As shown in Table~\ref{tab:compare_with_sota}, training the simplified versions of Maria or removing any visual signals from input components leads to worse performance in terms of relevance and diversity. 
In particular, the results on the ablation study validate that: 
(1) The performance improvement of dialog generation benefits from the MCP's effectiveness in aligning the representations of text and vision; 
(2) When training Maria, introducing VKB can further improve the quality and diversity of generated responses; 
(3) Rich visual knowledge, \textit{i.e.}, image region features and visual concepts, play a significant role in improving the performance of dialog generation. 
Especially, removing the visual concepts leads to a dramatic performance drop on diversity. 
The phenomenon is due to the lack of necessary visual concepts, Maria can not well understand the visual world knowledge when only learning from the visual features.

\begin{figure*}[t]
	\centering
	\includegraphics[width=0.8\textwidth]{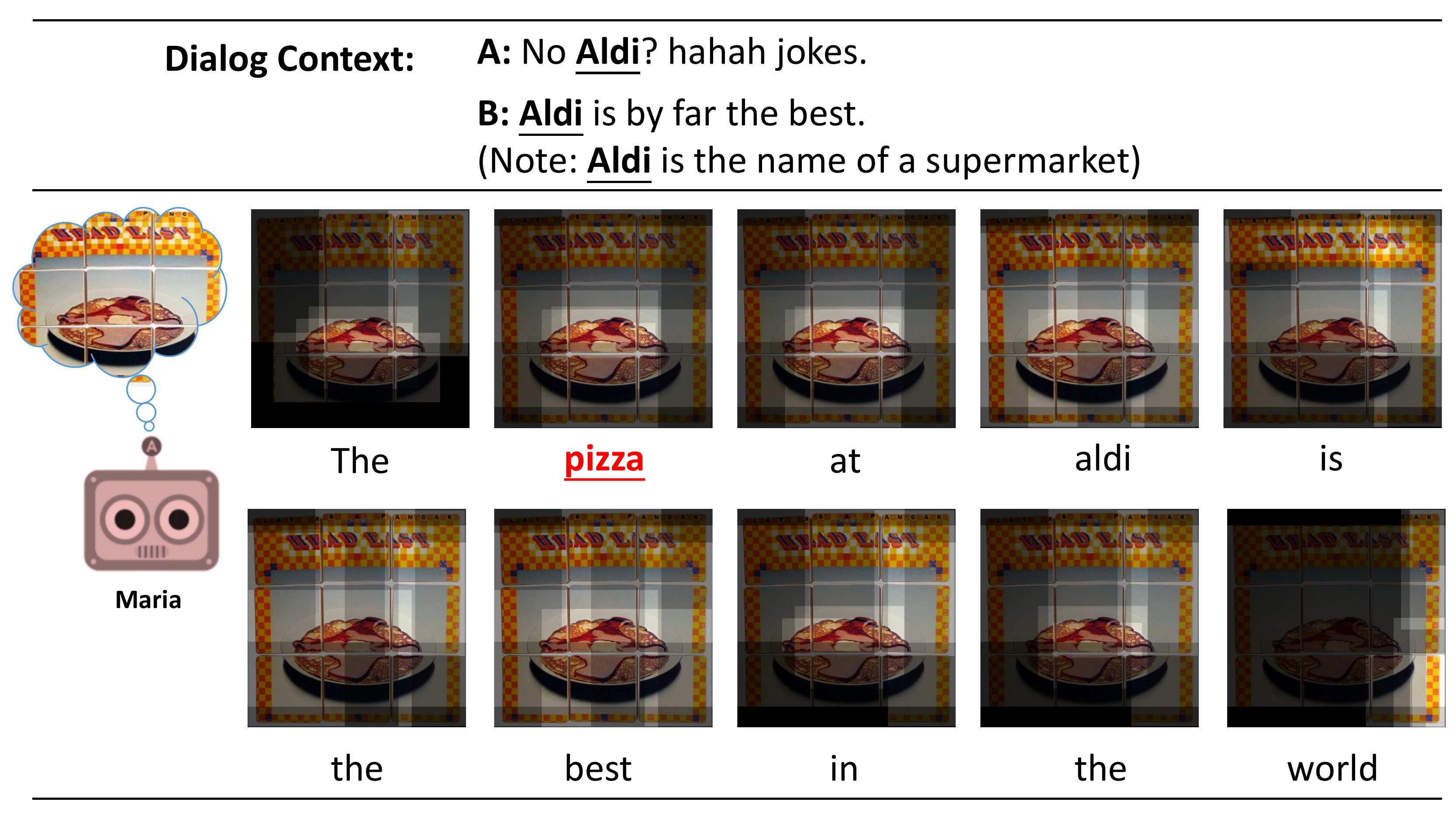}
	\caption{The visualization of attention weights on the retrieved image by Maria for an example.}
	\vspace{-1em}
	\label{fig:heatmap_visualization}
\end{figure*}

\subsection{Case Analysis}
\label{sec:case_analysis}

To further investigate the quality of responses generated by Maria, we put an example of generated responses in Figure~\ref{fig:heatmap_visualization}. 
As we can see from Figure~\ref{fig:heatmap_visualization}, when the context talks about the supermarket ``Aldi'', Maria can retrieve a ``pizza'' related image and generate the informative response grounded on it, \textit{i.e.}, ``the pizza at Aldi is the best in the world''.
This implies the commonsense that the supermarket usually has the pizza to sell.
It is also observed that Maria pays more attention to the relevant image regions when generating the word “pizza", which demonstrates that Maria could capture useful visual knowledge from the image and subsequently leverage it to generate commonsense-aware responses.
More cases are demonstrated in Appendices.

% \input{related}
% \vspace{-1.0em}
\section{Conclusions}

In this paper, we present Maria, a neural conversational agent powered by the visual world experiences.
It is able to retrieve the visual world experiences with users and generate human-like responses with some visual commonsense.
Extensive experiments demonstrate Maria achieves substantial improvements over the state-of-the-art methods in automatic and human evaluation.
The future works could include:
(1) Design a more precise and comprehensive image retriever to include multiple retrieval images;
(2) Combining the retrieve module and dialog generation into an end-to-end model, instead of learning them individually;
(3) Explore more efficient neural architectures to inject the visual knowledge into response generation.

\vspace{-1em}
\bibliographystyle{acl_natbib}
\bibliography{anthology,acl2021}

%\appendix
\begin{figure}[b]
	\centering
	\includegraphics[width=1\textwidth]{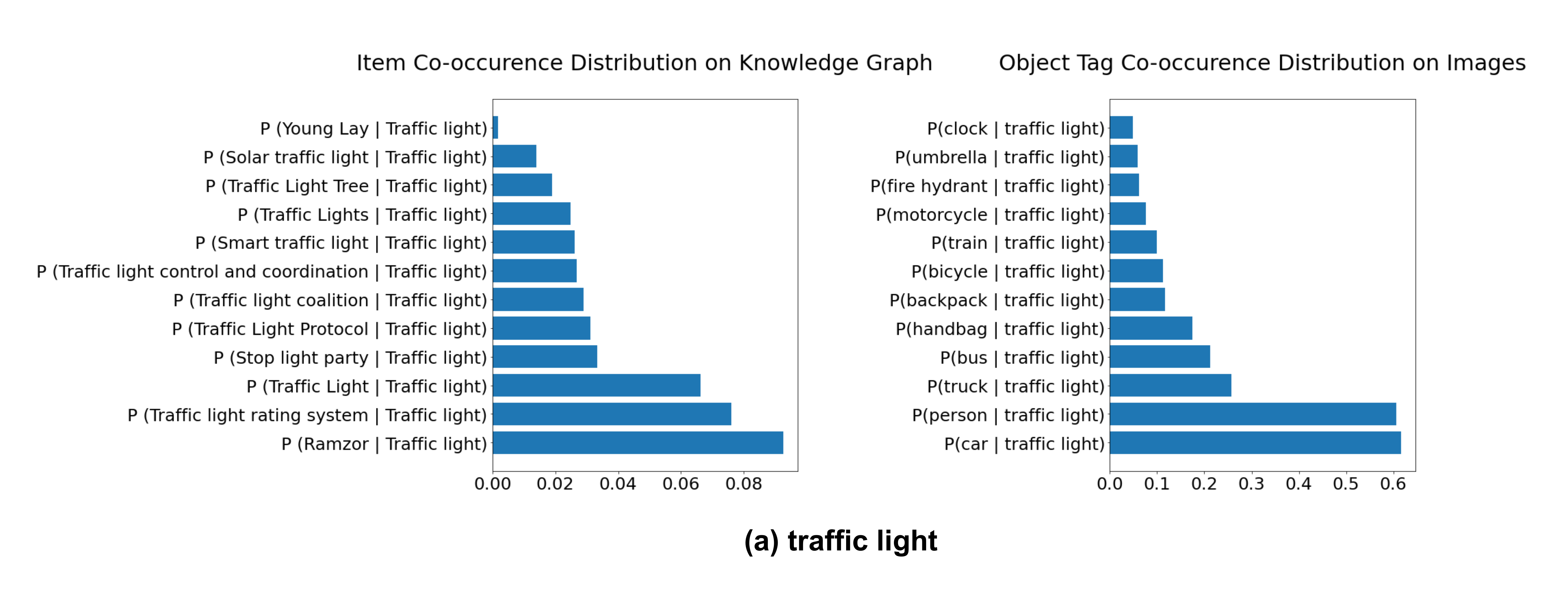}
	\label{fig:appendix_figure_distribution_1}
\end{figure}

\appendix
\section{Appendices}
In this section, we show more examples of word co-occurrence distributions on Google knowledge graph and MS-COCO images. Besides, some conversation samples produced by Maria and the baselines are also presented in Section~\ref{sec:more_case_analysis}.

\subsection{Word Co-occurrence Distribution Examples}
\label{sec:distribution_examples}

\begin{figure*}[t]
	\centering
	\includegraphics[width=1\textwidth]{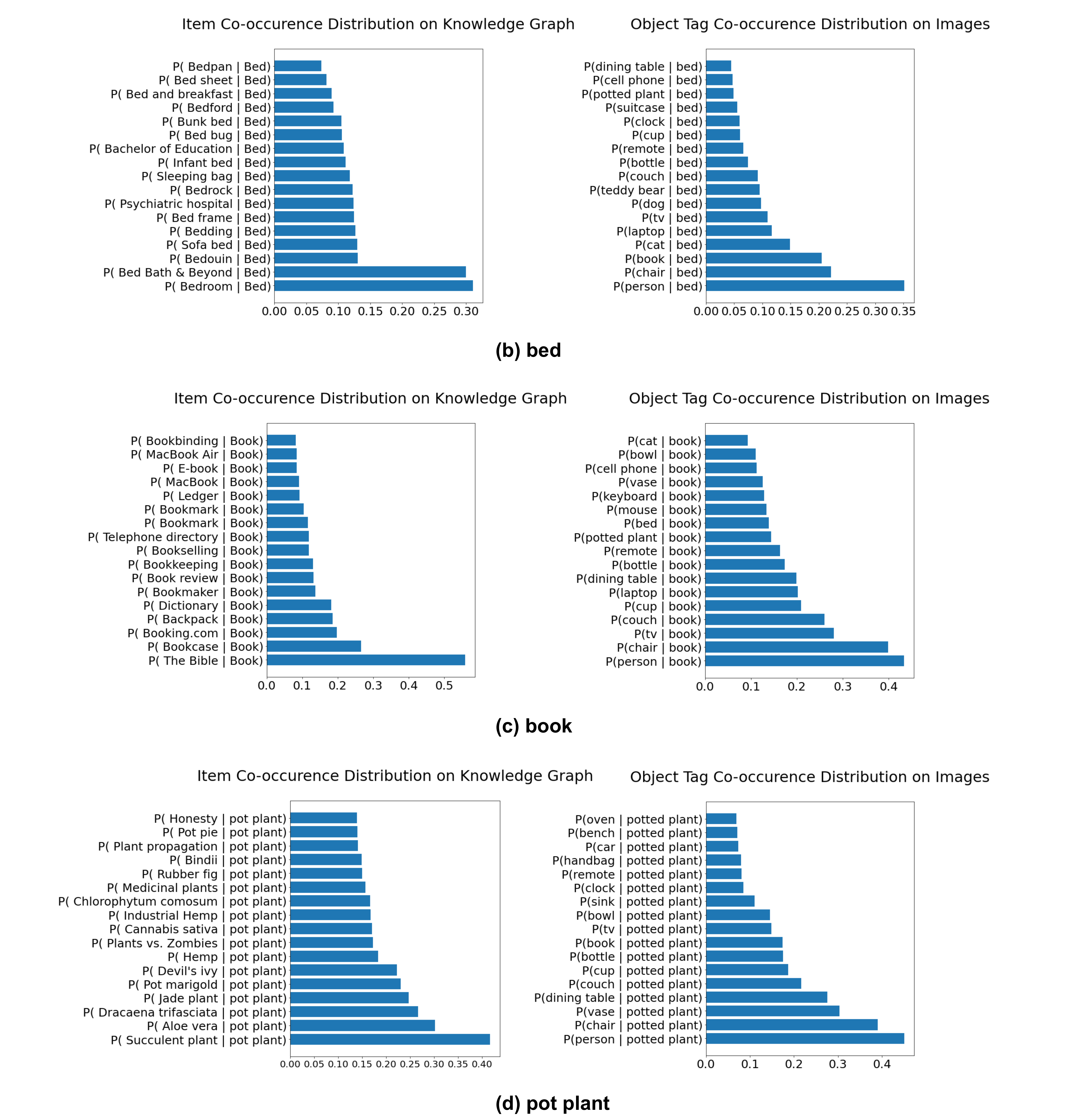}
	\caption{Supplementary examples of the word co-occurrence distribution on Google knowledge graph and MS-COCO images.}
	\label{fig:appendix_figure_distribution}
\end{figure*}

In Figure~\ref{fig:appendix_figure_distribution}, we present some supplementary examples of the word co-occurrence distribution on Google knowledge graph and MS-COCO images, including ``traffic light'', ``bed'', ``book'', and ``pot plant''. 
Figure~\ref{fig:appendix_figure_distribution} (a) shows the co-occurrence distributions of ``traffic light'' and other words on knowledge graph and images, respectively. 
As we can see, most of the co-occurred words with ``traffic light'' are the related concepts such as ``smart traffic light'', ``traffic light protocol'', ``traffic light rating system'', etc.
While the co-occurred words on images are usually ``car'', ``person'', ``truck'', ``bus'', etc, which we often see when walking by the traffic lights.
Interestingly, we found ``umbrella'' and ``clock'' also co-occurs with ``traffic light'' in some images.
For the former, the picture we can imagine is that people were holding the ``umbrellas'' when they walked through a zebra crossing under the ``traffic light''.
For the latter, the possible picture is that we can see both the ``traffic light'' and the ``clock'' on the top of a high building from a certain angle when walking on the street.
Similar observations can be also seen in other examples.
\vfill\eject
Most of the co-occurrence words on knowledge graph are logically-related concepts.
However, the co-occurrence relationship of object tags on images reflects some commonsense of our physical world, which implies some pictures that we human could easily imagine. 
This kind of knowledge is unique and inherent in images, but it can hardly be captured in the traditional knowledge bases, such as knowledge graph.

\subsection{Case Analysis}
\label{sec:more_case_analysis}

\begin{figure*}[t]
	\centering

    \includegraphics[width=\textwidth]{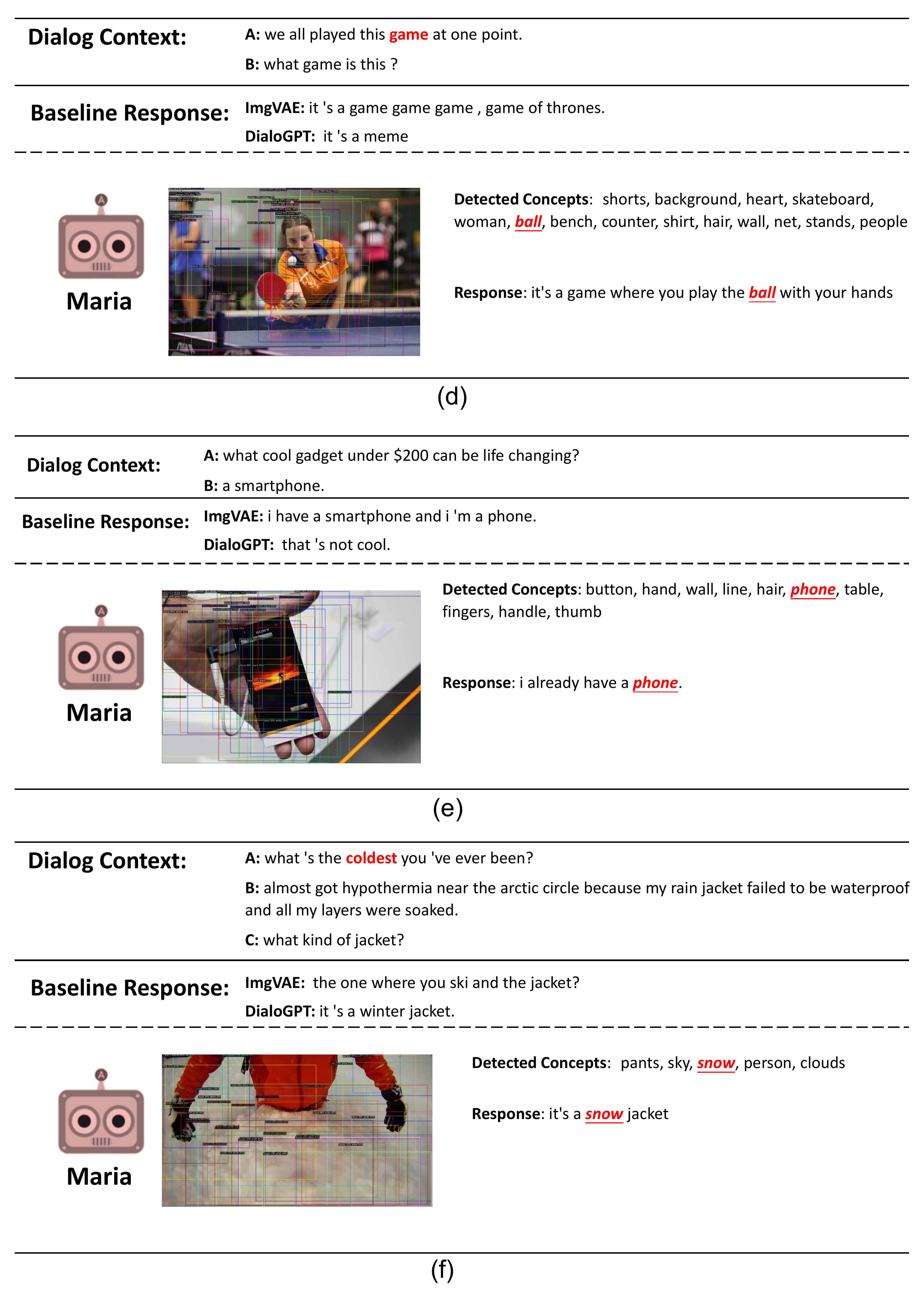}

    \caption{Case Study on the Reddit data from test split.}

	\label{fig:appendix_case}
\end{figure*}

Figure~\ref{fig:appendix_case} shows some cases from the test set of Reddit data.
We observe that the responses generated by Maria are more commonsensical and vivid than those of the baseline methods, which is consistent with our automatic and human evaluation results.
Interestingly, Maria is able to retrieve correlated images using the dialog contexts, which makes its response more human-like. 
For instance, case (a) shows that when the dialog context marvels at ``the pass of the world cup'', Maria recalls a football player and compliments him ``the best player in the world'';
case (b) shows that when the dialog context chats about the ``Canada weather'', Maria is aware of the fact that ``Canada'' is often ``snowy'' and then talks about ``Canada'' in a funny tone, ``I've never been to a place that doesn't have snow'';
case (c) shows that Maria understands that ``swan'' is sometimes ``dangerous'' when they are on the ``beach'';
case (d) shows that when the dialog context tries to guess one type of game, Maria recalls a ping-pong ``ball'' game and describes it;
and etc.

\begin{figure*}[t]
	\centering
    \includegraphics[width=\textwidth]{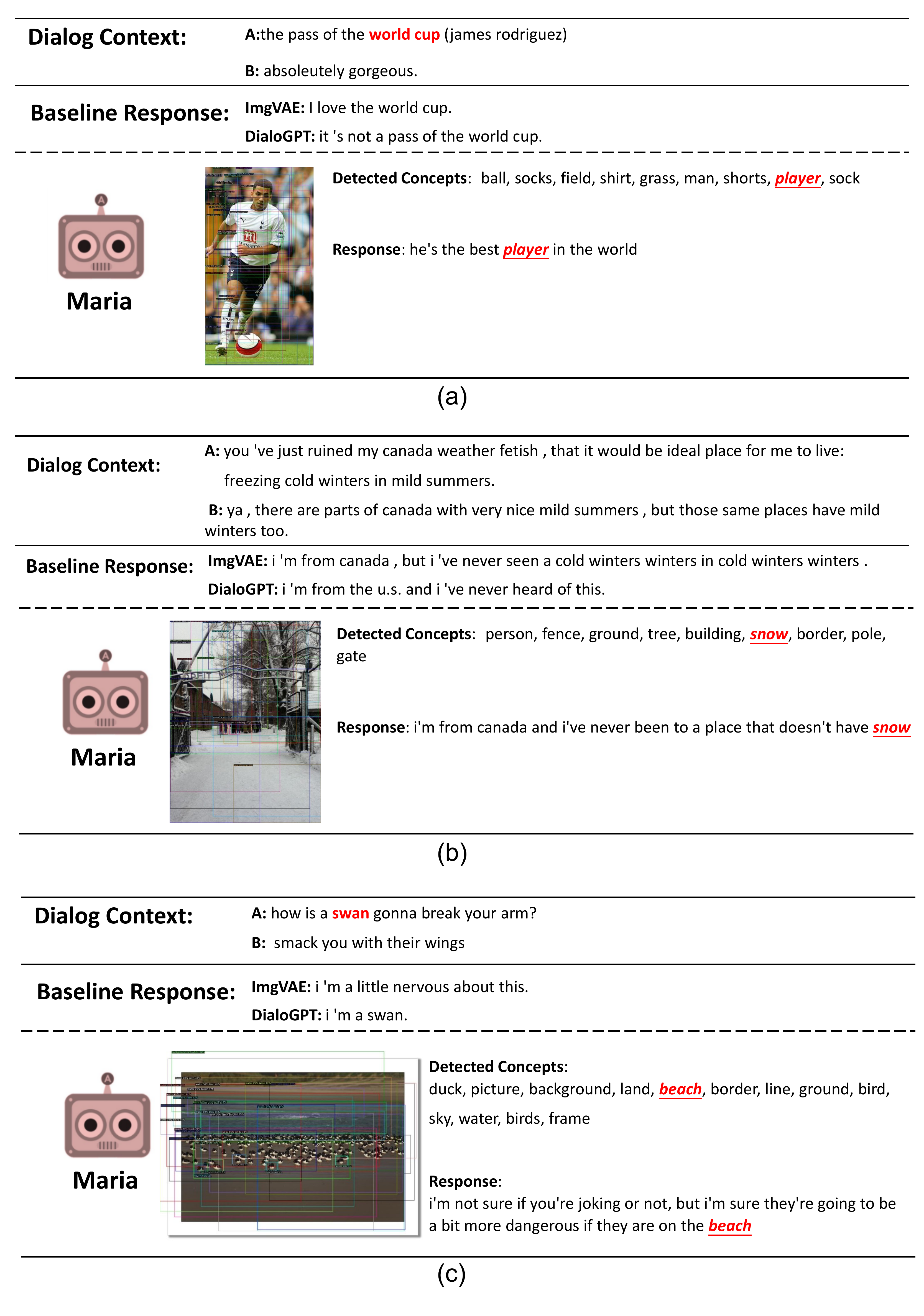}
\end{figure*}

\end{document}